\documentclass[10pt,twocolumn,letterpaper]{article}

\usepackage[pagenumbers]{cvpr} 

\usepackage{graphicx}
\usepackage{amsmath}
\usepackage{amssymb}
\usepackage{bbold}
\usepackage{booktabs}

\usepackage[accsupp]{axessibility}
\usepackage{adjustbox}

\usepackage[pagebackref,breaklinks,colorlinks]{hyperref}

\usepackage[capitalize]{cleveref}
\crefname{section}{Sec.}{Secs.}
\Crefname{section}{Section}{Sections}
\Crefname{table}{Table}{Tables}
\crefname{table}{Tab.}{Tabs.}

\usepackage{algorithm2e}
\RestyleAlgo{ruled}

\def\cvprPaperID{9432} 
\def\confName{CVPR}
\def\confYear{2022}

\begin{document}

\title{Use All The Labels: A Hierarchical Multi-Label Contrastive Learning Framework}

\author{Shu Zhang \quad Ran Xu \quad Caiming Xiong \quad Chetan Ramaiah\\
Salesforce Research\\
{\tt\small \{shu.zhang,ran.xu,cxiong,cramaiah\}@salesforce.com}
}
\maketitle

\begin{abstract}
Current contrastive learning frameworks focus on leveraging a single supervisory signal to learn representations, which limits the  efficacy on  unseen data  and  downstream  tasks. In this paper, we present a hierarchical multi-label representation learning framework that can leverage all available labels and  preserve the hierarchical relationship between classes. We introduce novel hierarchy preserving losses, which jointly apply a hierarchical penalty to the contrastive loss, and enforce the hierarchy constraint. The loss function is data driven and automatically adapts to arbitrary multi-label structures. Experiments on several datasets show that our relationship-preserving embedding performs well on a variety of tasks and outperform the baseline supervised and self-supervised approaches. Code is available at \url{https://github.com/salesforce/hierarchicalContrastiveLearning}.
\end{abstract}

\section{Introduction}

In the real world, hierarchical multi-labels occur naturally and frequently. Biological classification of organisms is structured in a taxonomic hierarchy. In e-commerce websites, retail spaces and grocery stores, products are organized  by several levels of categories. The hierarchical representation is a natural categorization of classes, and serves to efficiently represent the relationship between different classes. However, this relationship is seldom utilized in learning tasks, with traditional supervised approaches preferring to organize their classes in a flat list. In single task learning problems, where a model is learned for one objective only, a flat list of classes is a reasonable approach. However, in representation learning frameworks, where a single embedding function can be used in a variety of downstream tasks, utilizing all of the supervisory signal available is vital. In order to generalize to unknown downstream tasks and unseen data, the embedding function must represent the data concisely and accurately, which includes preserving the hierarchical categorization in the embedding space. 

However, representation learning  approaches that exploit this hierarchical relationship between labels have received very little attention. In recent years, several unsupervised \cite{he2020momentum, chen2020simple,Hjelm2019learningdeep} and supervised \cite{khosla2020supervised, ho2020exploit, schroff2015facenet, Weinberger2009triplet} metric learning frameworks have been proposed. These approaches typically rely on minimizing the distance between representations of a positive pair and maximizing the distance between negative pairs. In the unsupervised (self-supervised) setting \cite{wu2018unsupervised, tian2019contrastive, he2020momentum, chen2020simple}, the positive pairs are different views of the same image, most typically obtained by random augmentations of the anchor image \cite{chen2020simple, he2020momentum, tian2019contrastive}. In the supervised setting, labels are used to construct a wider variety of positive pairs, from different images of the same class and their augmentations \cite{ho2020exploit, khosla2020supervised}. Positive pairs constructed from augmentations of the anchor image, and pairs constructed from the anchor image and other images of the same class are considered to be equivalent, and the learning process attempts to minimize the distance between images in all of these positive pairs to the same degree.  While representations learned in this paradigm may be satisfactory for a downstream task based on the supervisory label such as category prediction, other tasks such as instance prediction, retrieval, attribute prediction and clustering can suffer due to the absence of direct supervision for these tasks. Additionally, these approaches do not support multi-label learning and are unable to utilize information about the relationship between labels. 

\begin{figure*}
\begin{center}
\includegraphics[ height=0.3\textheight]{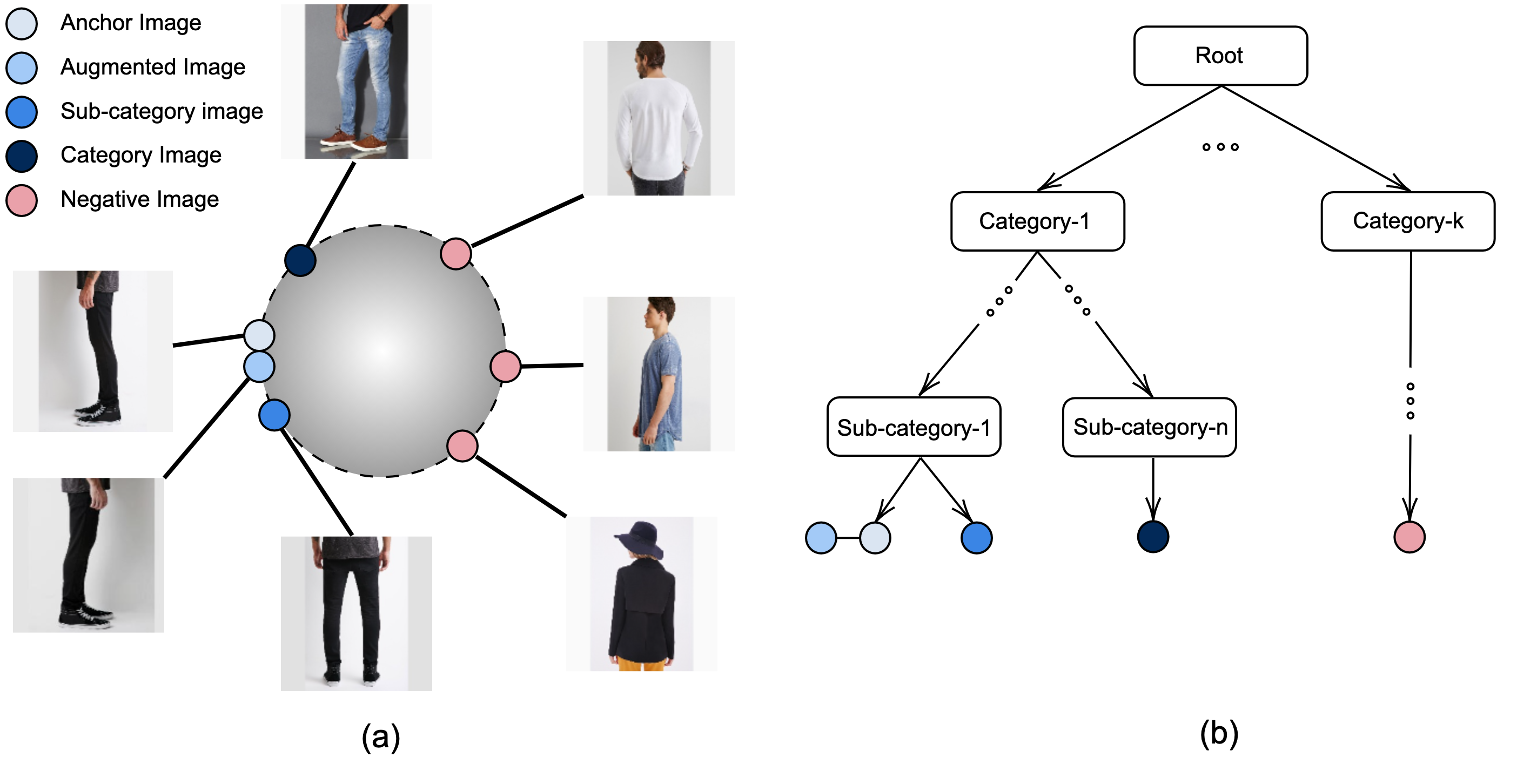}
\end{center}
   \caption{Hierarchical multi-label contrastive learning overview. A positive pair is constructed by pairing the anchor image with images drawn from all levels in the hierarchy. The learning objective of this work is to force positive pairs closer together, but the magnitude of the force is dependent on the common ancestry of the pair's labels. (a) The anchor image and corresponding positive pairings (blue) and negative pairings(red), visualized on a unit sphere. Different shades of blue nodes indicate their relationship to the anchor image, with darkness in shades of blue corresponding to increasing distance (both in the label space and representation space) from the anchor image. The red data points are from different categories in the dataset and hence form negative pairs with the anchor image. (b) Representation of the sample images from (a) in the label hierarchy. A tree data structure is used to to visualize the multi-labels.}
\label{fig:intro}
\end{figure*}

Formally, in the hierarchical multi-label setting, each data point has multiple dependent labels, and the relationship between labels is best represented in a hierarchy. See Figure \ref{fig:intro}(b) for a sample representation in a tree structure. For example, in the DeepFashion dataset\cite{Liu2016deepfashion}, each data point has 3 hierarchically structured labels: category (Denim, Cardigan, Shirts etc), product (identified by product id) and variation (typically color / pattern variations). For the anchor image in Figure \ref{fig:intro}(a),  which belongs to a specific product in the Denim category , the sub-category image is a different sample from the same product, and the category image is from a different product in the same category. All the negative images are from different categories. 

Our proposed approach leverages all the available labels to learn an embedding function that can preserve the label hierarchy in the embedding space. We develop a general representation learning framework that can utilize available ground truth and learn embeddings that generalize to a variety of downstream tasks. We present two novel losses (and their combination) that exploit the relationship between hierarchical multi-labels and learn representations that can retain the label relationship in the representation space. The Hierarchical Multi-label Contrastive Loss (HiMulCon) enforces a penalty that is dependent on the proximity between the anchor image and the matching image in the label space. In this setting, we define proximity in the label space as the overlap in ancestry in the tree structure. The Hierarchical Constraint Enforcing Loss (HiConE) prevents the hierarchy violation, that is, it ensures that the loss from pairs farther apart in the label space is never smaller than the loss from pairs that are closer. Models learned under this framework can be used exactly like traditional representation learning frameworks, a model is trained with our novel loss functions to learn an efficient encoder network, and embeddings generated from this approach can be used in a variety of downstream tasks.

Our framework is not limited to the hierarchical multi-label scenario. It reduces to the supervised contrastive approach \cite{khosla2020supervised} when only single level labels are available, and to the SimCLR \cite{chen2020simple} approach when no labels are available. We demonstrate the efficacy of our framework in comparison to Khosla \etal \cite{khosla2020supervised}, Chen \etal \cite{chen2020simple} and a standard cross entropy based approach on downstream tasks such as category prediction, sub-category retrieval and clustering NMI \cite{vinh2010information}. These tasks also show that our approach preserves the hierarchical relationship between labels in the representation space, and generalizes to unseen data as well.

\begin{figure*}
\begin{center}
\includegraphics[width=0.85\textwidth]{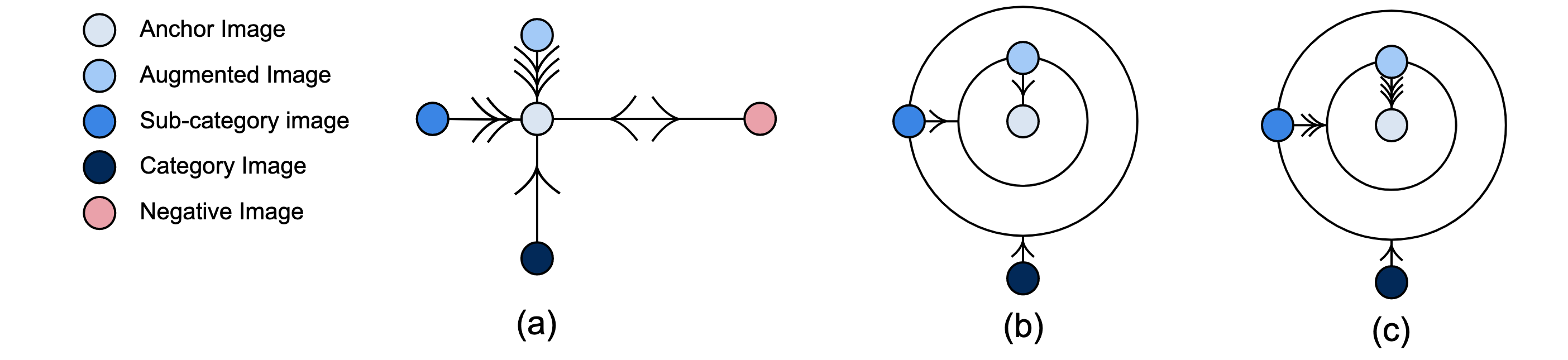}
\end{center}
   \caption{Conceptual visualization of the losses.(a) HiMulCon loss: A penalty, proportional to the proximity in the label space, is enforced on each positive pair. Images that are closer to the anchor image in the label space have a higher penalty (represented by the arrows), forcing them closer together. (b) HiConE loss: The hierarchy constraint is enforced by ensuring that image pairs that are farther away in the label space will never have a lower loss than image pairs that are closer.  (c) Combined (HiMulConE) loss: (a) and (b) are combined, the penalty is applied in combination with the hierarchy preserving constraint. The loss term w.r.t negative samples is unchanged, and the negative images are omitted in this example.}
\label{fig:concept}
\end{figure*}

\section{Related Work}
Contrastive learning was first studied in the self-supervised setting \cite{wu2018unsupervised, tian2019contrastive, he2020momentum, chen2020simple, Hjelm2019learningdeep, Tschannen2020Mutual}, typically relying on a pretext task to learn embeddings. The supervision for the pretext tasks would typically be generated from the data itself. Examples include denoising \cite{vincent2008extracting}, colorization \cite{zhang2017split}, image recognition \cite{Henaff2020dataefficient}, object detection and image segmentation \cite{Wang2020Dense}, action recognition \cite{Pan2021Videomoco} and patch ordering \cite{doersch2015unsupervised}. The method in van den Oord \etal \cite{Oord2018representation} used a probabilistic contrastive loss to capture mutual information between different views of data. They showed the effectiveness of their method on four domains: speech, images, text and reinforcement learning. Li \etal \cite{Li2020prototypical} performed clustering to find prototypical embedding, which is similar to an image embedding. The efficacy of learning representations in a contrastive fashion led to the development of supervised constrastive approaches. Khosla \etal \cite{khosla2020supervised} extended Chen \etal \cite{chen2020simple} to the supervised setting. The supervised contrastive learning approach formulated positive pairs by sampling from different instances of the same class, as opposed to augmenting different views of the same image in the unsupervised setting. They show that the supervised contrastive learning is a generalization of the triplet loss \cite{Weinberger2009triplet} and the n-pair loss objectives \cite{Sohn2016Npair}. Ho \etal \cite{ho2020exploit} introduced an approach to utilize multiple instances of an object interchangeably to learn a viewpoint invariant representation in a self-supervised and semi-supervised setting.

Ma{\l}ki{\'n}ski \etal \cite{malkinski2020multi} proposed a contrastive learning framework for visual reasoning in the multi-label setting, but under the assumption that any pair of images can be considered to be positive if they share even one label. Zhao \etal \cite{Zhao2011largescale} proposed a normalized class-similarity-weighted sums for classification in the hierarchical setting, which is a replacement of the logistic regression. Wu \etal \cite{Wu2016Learning} learned a hierarchical classifier by calculating each hierarchy level's probability. Cho \etal \cite{Cho2019Leveraging} designed a fashion hierarchical classification model to classify fashion images. Bilal \etal\cite{Bilal2017Doconvolutional} and Yan \etal \cite{Yan2015HDCNN} designed a deep network branch on side of a general network, and strategically learn super-classes at each hierarchical level. Instead of solving the label imbalance problem, \cite{Bertinetto2020Bettermistakes} designed two cross entropy variations to handle top-k errors. 
Wehrmann \etal \cite{wehrmann2018hierarchical} introduced a multi-step pipeline  learning a network which output a prediction for each layer in the hierarchy. 
Giunchiglia \etal \cite{giunchiglia2020coherent} introduced the coherent hierarchical multi-label classification networks, which enforced the hierarchy constraint (described in Section \ref{sec:HiConE}). They introduced a modified version of binary cross entropy loss, where a separate module would modify the model confidence such that the confidence associated with all classes in the hierarchy will be equal or higher than the lowest class. Wang \etal \cite{Wang2019Rankedlist} proposed a structured loss based on the similarities between query and gallery features. The method in \cite{Kim2021} was built on Triplet loss, where levels of features are defined as degrees of similarities between pairwise points. These methods are fundamentally different from the proposed method, which directly builds the hierarchy on the natural characteristics of the data.

Ge \etal \cite{ge2018deep} constructed a hierarchical triplet loss, wherein the hierarchical tree representation is constructed using a single-level label structure by utilizing the intraclass distances to formulate a grouping mechanism, which is then used for hard negative mining and in the loss function. This approach differs from our approach in two important ways, first, it relies on the triplet loss instead of the contrastive loss. The triplet loss is a special case of the contrastive loss \cite{khosla2020supervised}. The second difference lies in the construction of the hierarchical tree. Our approach relies on the multi-level label information to construct the tree, whereas Ge \etal \cite{ge2018deep} constructs it on the fly from the data itself. The formulation of the tree from the data can result in propagating biases in the underlying model and the data to the representation learning framework and can be prone to noise. A recent method that was proposed by Garnot \etal \cite{Garnot2021} modeled the hierarchical class structure by integrating class distance into a prototypical network, where the use of a hierarchical tree is different in our approach.

Although our experiments include image classification and image retrieval, our goal is different from multi-task methods \cite{wehrmann2018hierarchical, giunchiglia2020coherent, Berman2019multigrain}. The proposed algorithm is a general hierarchical multi-label representation learning framework that can be applied to any downstream task. Our approach is agnostic to downstream tasks and is not directly optimized for them.

\section{Methodology}
In order to better explain our approach, we define some of the terminology that will be used throughout this work. A hierarchical multi-label dataset refers to a dataset where each data point has multiple related labels associated with it, and the dependency is best described in a directed acyclic graph or a tree. Leaf nodes represent a unique image identifier, and all non-leaf nodes in the tree represent label information at various levels. The levels are analogous to depth in a tree structure. The lower levels correspond to broader categories (closer to the root of the tree), with the lowest level corresponding to the category label. For example, in Figure \ref{fig:intro}, the ``DENIM" category would be the lowest label for the anchor image, and \textit{sub-category-1} would be the highest level label. Leaf nodes would correspond to image identifiers. Positive pairs at a level $l \in L$ are formed by identifying a pair of images that have common ancestry up to level $l$ and diverge thereafter. Referring again to Figure \ref{fig:intro}, the anchor image and the category image form a pair at the category level, as they only have the category label to be common between them. In graph terminology, a pair of images at level $l$ implies that they will have their lowest common ancestor at level $l$.

Our method is constructed similar to the supervised contrastive learning \cite{khosla2020supervised} approach. First, an encoder network and a projection head is learned using all of the available hierarchical labels. The encoder networks weights are then frozen, and no finetuning is done on the encoder network for downstream tasks. If additional learning is necessary for downstream tasks, in classification tasks for example, a seperate classifier is trained with the embeddings generated by the encoder acting as the input to the classifier.


\subsection{Contrastive Learning} \label{sec:InfoNCE}

The contrastive learning loss introduced in Chen \etal \cite{chen2020simple}, was originally a self-supervised learning method. Such methods can pull an anchor and its augmented version together in the embedding space, while the anchors and negative samples are pushed apart. A set of $N$ randomly sampled labeled pairs is defined as  $\{x_k, y_k\}$, where $x$ and $y$ represent the samples and labels individually and $k=1, ..., N$. Two augmentations are applied to each sample. Let $i$ be the index of one augmented sample, and $j$ be the index of the other, where $i \in A = \{1, ..., 2N \}$ and $j \neq i$. $i$ is the anchor and $j$ is the positive sample. The contrastive loss is defined as 
\begin{equation}
    L^{\mathrm{self}} =  - \sum_{i\in A}\frac{\exp(f_{i}\cdot f_{j}/\tau )} {\sum_{k \in A \backslash i} \exp(f_{i}\cdot f_{k}/\tau )}
\label{eq:selfsup}
\end{equation}
Here, $f$ represents the feature vector in the embedding space, and $\tau$ is a temperature parameter. Intuitively, the numerator calculates the inner dot product between an anchor $i$ and its positive sample $j$. The denominator calculates all the inner dot products between $i$ and all negative samples, where totally $2N - 1$ samples are calculated.

The supervised contrastive learning \cite{khosla2020supervised} extends Eq. \ref{eq:selfsup} to a supervised scenario. Particularly, given the presence of labels, positive pairings for the anchor goes from one-to-many positive-negative samples in SimCLR \cite{chen2020simple}, to many-to-many samples. The loss is defined as 
\begin{equation}
    L^{\mathrm{sup}} =  \sum_{i\in I} \frac{-1}{\left | P(i) \right |} \sum_{p\in P} \log \frac{\exp(f_{i}\cdot f_{p}/\tau )} {\sum_{a \in A \backslash i} \exp(f_{i}\cdot f_{a}/\tau )}
\label{eq:SupCon}
\end{equation}
$P$ represents the indices of all positives in the multi-view batches except for $i$. $A$ represents all images in the batch, and $a \in A \backslash i$ is all images in the batch except the $ith$ image. Therefore, the supervised contrastive loss consolidates information of all the positive samples in the numerator, and can essentially exploit the contrastive power between positive and negative samples.

\subsection{Hierarchical Multi-label Contrastive Loss}

Although the supervised contrastive learning in Eq. \ref{eq:SupCon} can distinguish between multiple positive pairs, it is only designed for single labels. Define $L$ as the set of all label levels, and $l \in L$ is a level in the  multi-label. Then the loss for a pairing of the anchor image, indexed by $i$ and a positive image at level $l$ is defined as

\begin{equation}
L^{\mathrm{pair}}(i,p_l^i) = \log \frac{\exp(f_{i}\cdot f_{p}^l/\tau )} {\sum_{a \in A \backslash i} \exp(f_{i}\cdot f_{a}/\tau )}
\end{equation}

The hierarchical multi-label contrastive loss (HiMulCon) can then be defined as:

\begin{equation}
    \begin{aligned}
    L^{\mathrm{HMC}} = \sum_{l \in L} \frac{1} {\left | L \right |} \sum_{i\in I} \frac{- \lambda_l}{\left | P_l(i) \right |} \sum_{p_l\in P_l} L^{\mathrm{pair}}(i,p_l^i)
    \end{aligned}
\label{eq:HiMulCon}
\end{equation}
where $\lambda_l= F(l)$ is a controlling parameter that applies a fixed penalty for each level in the hierarchy, $P_l$ is the set of positive images for anchor image indexed by $i$. $F$ is heuristically chosen and scales with $l$. See supplementary material for a study on different choices for $F$. Figure \ref{fig:concept}(a) provides a conceptual illustration of this loss. 

The HiMulCon applies higher penalties to image pairs constructed from higher levels in the hierarchy, forcing them closer than pairs constructed from lower levels in the hierarchy. Note the construction of the loss with regards to the interaction between pairs at different levels. Pairs formed at the highest level will have all other paired images from lower levels form negative pairs at the higher levels, and the negative pairs formed by pairs with some level of lower common ancestry naturally form hard negative samples, hence becoming a form of hard negative mining. In addition, the $\lambda_l$ term contributes to preserving the hierarchy explicitly.  

If there is only one-level label, the HiMulCon loss reduces to the supervised contrastive loss. The supervised contrastive loss is therefore a special case of the HiMulCon.

\subsection{Hierarchical Constraint Enforcing Loss} \label{sec:HiConE}

The Hierarchical Constraint Enforcing Loss, HiConE, enforces the hierarchical constraint in the representation learning setting. In the classification setting, as described in Wehrmann\etal \cite{wehrmann2018hierarchical} and  Giunchiglia \etal \cite{giunchiglia2020coherent}, the hierarchical constraint ensures that if a data point belongs to a class, it should also belong to its ancestors. This can be defined in terms of confidence scores, where a class higher in the hierarchy cannot have a lower confidence score than a class lower in the ancestry sequence. Adapted to the contrastive learning scenario, we define the hierarchical constraint as the requirement that the loss between image pairs from a higher level in the hierarchy will never be higher than the loss between pairs from a lower level. This observation leads us to develop an hierarchical constraint enforcing loss (HiConE).

If we define $L^{\mathrm{pair}}_{\mathrm{max}}$ as the maximum loss from all positive pairs at level $l$:

\begin{equation}
L^{\mathrm{pair}}_{\mathrm{max}}(l) = \max_{(i,p_l^i)} L^{\mathrm{pair}}(i,p_l^i)
\end{equation}

Then, the HiConE loss $L^{\mathrm{HCE}}$ is defined as

\begin{equation}
    \begin{aligned}
    \sum_{l \in L} \frac{1} {\left | L \right |} \sum_{i\in I} \frac{- 1}{\left | P(i) \right |} \sum_{p_l\in P_l}  {\max}(L^{\mathrm{pair}}(i,p_l^i), L^{\mathrm{pair}}_{\mathrm{max}}(l-1))
    \end{aligned}
\label{eq:HiConE}
\end{equation}
HiConE is computed sequentially in decreasing order of $l$ from $L$ to $0$, which helps ensure that the pair loss at level $l-1$ can never be less than the max pair loss at $l$.  Figure \ref{fig:concept}(b) has a conceptual visualization for this loss, where pairs formed at lower levels in the hierarchy will never have a higher loss than pairs formed at a lower level in the hierarchy. 

\subsection{Hierarchical Multi-label Constraint Enforcing Contrastive Loss}

Intuitively Eq. \ref{eq:HiMulCon} acts as an independent penalty defined on the level, whereas Eq. \ref{eq:HiConE} is a dependent penalty that is defined in relation to the losses computed at the lower levels. We can combine the two losses to form the combined loss, the Hierarchical Multi-label Constraint Enforcing Contrastive Loss (HiMulConE), $L^{\mathrm{HMCE}}$

\begin{equation}
    \begin{aligned}
    \sum_{l \in L} \frac{1} {\left | L \right |} \sum_{i\in I} \frac{- \lambda_l}{\left | P(i) \right |} \sum_{p_l\in P_l}  {\max}(L^{\mathrm{pair}}(i,p_l^i), L^{\mathrm{pair}}_{\mathrm{max}}(l-1))
    \end{aligned}
\label{eq:combined}
\end{equation}

Note that the combined loss is essentially adding the $\lambda_l$ term to Eq. \ref{eq:HiConE}, giving us a loss term that has a level penalty as well as the hierarchy constraint enforcing term.
\subsection{Hierarchical Batch Sampling Strategy}
\label{sec:sampling}
Wu \etal \cite{wu2017sampling} highlighted the importance of sampling in representation learning. Khosla \etal \cite{khosla2020supervised} also showed that having a large number of hard positives/negatives in a batch leads to improved performance. In a hierarchical multi-label setting, it becomes necessary to ensure that each batch has sufficient representation from all levels of the hierarchy for each anchor image. Hence,we design a custom batch sampling strategy which ensures that each image can form a positive pair with images that share a common ancestry at all levels in the structure. The approach is straightforward: randomly sample an anchor image and get the label hierarchy. For each label in the multi-label, randomly sample an image in the sub-tree such that the anchor image and the sampled image have common ancestry up to that label. Steps are taken to ensure that each image is sampled only once in an epoch.

For example, in Figure \ref{fig:intro} (b), the anchor image would be sampled. Positive pairings from each level need to be sampled next. First, a random image from sub-category-1 will be sampled. Next, a random image from category-1 but not sub-category-1 will be sampled. This process is repeated at all levels in the hierarchy. Finally, augmented versions of these images are also generated. Once completed, another anchor image is sampled randomly and the process repeats until $batch\_size$ number of images have been sampled.

\section{Experiments}

We evaluate our loss on three downstream tasks: image classification, image retrieval accuracy on sub-categories and NMI for clustering quality. We study the generalizability of our approach by evaluating the performance of our encoder network on unseen data. We also present qualititative results with t-SNE visualizations \cite{van2008visualizing}.

\begin{table*}
\begin{center}
\begin{tabular}{ c|c|c|c|c } 
    \toprule[2pt]
 & ImageNet\cite{ILSVRC15} & DeepFashion\cite{Liu2016deepfashion} & iNaturalist\cite{van2018inaturalist} & ModelNet40\cite{Wu2015shapenets} \\ 
 \toprule[2pt]
 SimCLR\cite{chen2020simple} & 69.53 &  70.38 & 54.02 & 79.26  \\
 
 Cross Entropy & 77.60 & 72.44 & 56.86 & 81.31 \\ 

 SupCon \cite{khosla2020supervised} & 78.70 & 72.82 & 57.28 & 81.60 \\ 
 
 Guided \cite{Garnot2021} & 76.60 & 72.61 & 57.33 & 83.49 \\

 HiMulConE (Ours) & \textbf{79.14} & \textbf{73.21} & \textbf{59.40} & \textbf{88.46} \\ 
 \bottomrule[2pt]
\end{tabular}
\end{center}
\caption{Top-1 classification accuracy on the full datasets. Standard datasets and splits as described in the original papers are used here. For ImageNet and iNaturalist datasets, the task is classification at the finest sub-category level, whereas super-category level classification is performed for DeepFashion and ModelNet40. All baseline results were reproduced.}
\label{table:classification_full}
\end{table*}

\begin{figure*}[t]
\begin{minipage}{0.33\linewidth}
\includegraphics[width=1\linewidth]{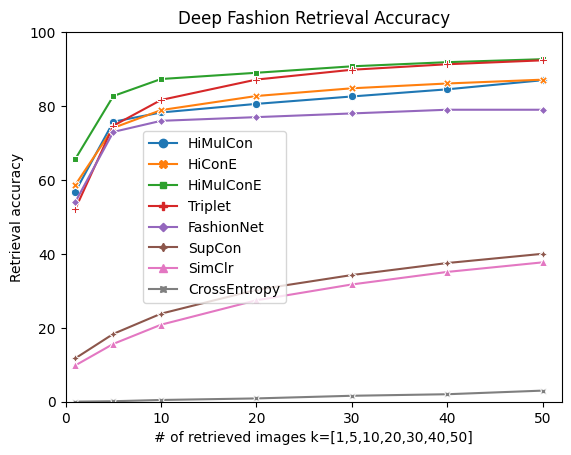}
\end{minipage}%
    \hfill
\begin{minipage}{0.33\linewidth}
\includegraphics[width=1\linewidth]{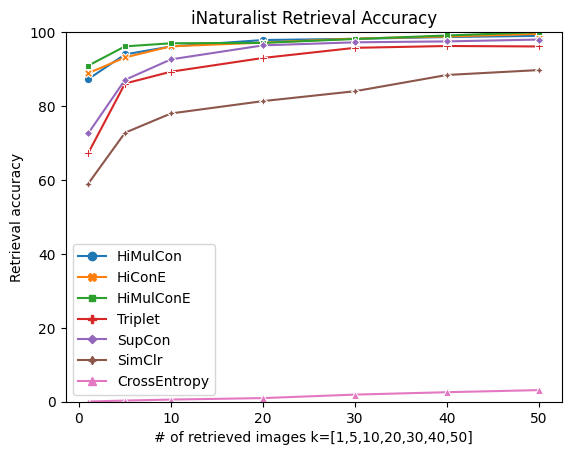}
\end{minipage}%
    \hfill
\begin{minipage}{0.33\linewidth}
\includegraphics[width=1\linewidth]{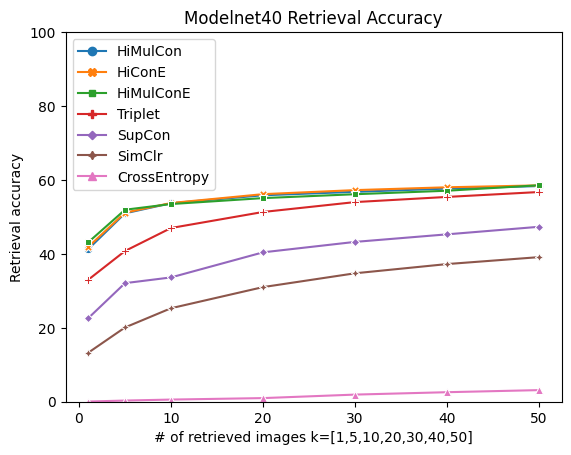}
\end{minipage}%
    \hfill
\caption{Retrieval results on the full datasets.}
\label{fig:fullretreival}
\end{figure*}


\subsection{Datasets}

We experiment with several popular datasets: ImageNet \cite{ILSVRC15}, DeepFashion In-Shop \cite{Liu2016deepfashion}, iNaturalist \cite{van2018inaturalist} and ModelNet40 \cite{Wu2015shapenets}. In order to showcase our results against popular benchmarks, we present results in the standard configurations of these datasets. We also split some of these datasets into seen and unseen sets, where we use the seen sets to train the encoder network, and evaluate performance of our approach and relevant baselines on the unseen dataset.  We use the split to show that proposed losses are able to learn generalized representations that work well on unseen data.

DeepFashion dataset is a large-scale clothing dataset with more than 800K images. We use the In-Shop subset in our experiments as it has three-level labels: category, product ID and variation. The variation can be different colors or sub-styles for the same product. ModelNet40 is a synthetic dataset of 3,183 CAD models from 40 object classes. It has two-level hierarchical labels: category and CAD image ID. iNaturalist is a species classification dataset, with two levels in the hierarchy, a super-category for the genus, and species categories. ImageNet classes are hierarchicaly structured using the WordNet \cite{miller1998wordnet} hierarchy, we use the label hierarchy published in the \textit{Robustness} library \cite{robustness}.

\SetKwComment{Comment}{/* }{ */}
\begin{algorithm}
\caption{Hierarchical Loss implementation}\label{alg:one}
\KwData{Batch labels $B$ of size N $\times$ L, Mask $M$ of size N $\times$ L, features $F$ of size N $\times$ d}
$l \gets N-1$\;
$M \gets \mathbb{1}$\;
$batchLoss \gets 0$\;
\While{$l \geq 0$}{
    $M[:,l] = 0$\;
    $levelMask = M \odot B$\;
    $i \gets 0$\;
    \texttt{\\}
    $levelPairings = torch.stack([torch.all($ \:\:\:\:\: $torch.eq(levelMask[i], levelMask), dim=1)$
                                       \:\:\:\:\:\: $for \;i\; in\; range(N)])$;\\
    \texttt{\\}                                   
    $levelLoss = loss(features, levelParings)$ \Comment*[r]{proposed loss functions}
    $batchLoss= batchLoss + levelLoss$\;
}
\end{algorithm}

\subsection{Implementation Details}
We adopt a pre-trained ResNet-50 \cite{He2016resnet}, which was trained on ImageNet \cite{deng2009imagenet}, as the encoder network. For the ImageNet experiments, we train the model from scratch. We finetune on our datasets for 100 epochs. Specifically, we finetuned the parameters of the fourth layer of the ResNet-50 as well as a multi-layer perceptron header \cite{khosla2020supervised} on the seen dataset with the proposed losses. The optimizer is SGD with momentum \cite{Ruder2016overview}. The encoder model weights are frozen after the network is trained in this fashion. We train an additional linear classifier for 40 epochs to obtain the classification accuracy. We use the same setup for all models. Although the proposed losses can be trained together with the linear classifier, we find that the performance gap is rather small. The batch size in our experiments is 512, and we use the temperature $\tau$ as 0.1 in all experiments. We start from the learning rate as 0.1, and decrease it by 10 for every 40 epochs. The augmentations are the same as \cite{khosla2020supervised,he2020momentum}.

The hierarchical relationship, and the batch sampling strategy described in section \ref{sec:sampling} requires a careful implementation of the loss calculation. Since a pairing can be a positive pair at a lower level and a negative pair at a higher level, the loss computation is aggregated by layers, with the highest layers calculated first. Restructuring the losses to be computed at each level and aggregated to form the batch loss allows for direct tensor operations and greatly speeds up the computation. See Algorithm \ref{alg:one} for a PyTorch\cite{NEURIPS2019_9015} based pseudo code. Starting from the highest level, a $N \times N$ pairing mask is constructed at each level where $levelPairing[i,j]=1$ $if\; labels[i]==labels[j] \;at\; level\; l\; else\; 0$. In the single-level scenario, $levelPairings$ is only computed once, and all images of the same class will be set, reducing to Khosla \etal \cite{khosla2020supervised}. The $levelLoss$ is defined in Eq. \ref{eq:HiMulCon}, \ref{eq:HiConE} and \ref{eq:combined} individually.  In the unlabeled scenario, $levelPairings$  will be a diagonal matrix, and this implementation reduces to SimCLR \cite{chen2020simple}. 

\begin{table}
\begin{center}
\begin{adjustbox}{width=0.9\columnwidth,center}
\begin{tabular}{ c|c|c|c|c } 
 \toprule[2pt]
    \multicolumn{1}{c|}{} & \multicolumn{2}{|c|}{DeepFashion} & \multicolumn{2}{|c}{ModelNet40} \\
 \toprule[1.5pt]
  & Seen & Unseen & Seen & Unseen \\ 
  \toprule[1.5pt]
 SimCLR & 70.26 &  68.12 & 77.09 & 72.26  \\
 
 Cross Entropy & 77.81 & 71.94 & 85.17 & 79.77 \\ 
 
 SupCon & \textbf{81.46} & 73.93 & 88.33 & 79.28 \\ 
 
  Guided & 79.34 & 74.04 & 89.01 & 82.22 \\
 
 HiMulCon & 80.54 & 74.88 & 89.28 & 84.44 \\ 
 
 HiConE & 80.67 & 75.28 & 89.09 & 84.40 \\ 
 
 HiMulConE & 80.52 & \textbf{75.29} & \textbf{89.45} & \textbf{85.37} \\ 
 \bottomrule[2pt]
\end{tabular}
\end{adjustbox}
\end{center}
\caption{Top-1 classification accuracy on the seen / unseen splits of DeepFashion In-Shop and ModelNet40. }
\label{table:classification_top1}
\end{table}
\subsection{Classification Accuracy}

We compare the proposed loss functions with SimCLR, an unsupervised contrastive loss \cite{chen2020simple}, two supervised learning losses functions: cross entropy and supervised contrastive loss (SupCon) \cite{khosla2020supervised}, and a state-of-the-art method (Guided) \cite{Garnot2021} that incorporates the idea of hierarchical labels. We do not compare with other metric learning approaches as Khosla \etal \cite{khosla2020supervised} showed that popular metric learning approaches like triplet loss \cite{schroff2015facenet, Weinberger2009triplet} are special cases of SupCon. The cross entropy uses flat list of labels and the softmax \cite{Bishop2006} function to train a classifier, and SupCon uses the labels to construct positive pairs in order to train a contrastive loss. The results are presented in Table \ref{table:classification_full}, and the classification results reported here are on the standard configurations of the datasets. The supervised approaches are very competitive for this task, as the encoder is also trained with the same supervisory signal as the classification task. Although our approach has access to additional labels during the representation learning phase, these labels are not used in training the classifier. All approaches evaluated here have exactly the same classifier training mechanism.


\subsection{Image Retrieval Accuracy}
\label{sec:retrieval}
This downstream task here is to retrieve images from the gallery that are the same class as the query image. The top-k accuracy is usually adopted to measure if a query image class can be found in the top-k retrieved results from the gallery. In this task, class is used to refer to the finest sub-category ID in the dataset. For DeepFashion dataset, the test set forms the query images, in ModelNet, we split the data into train, validation and test sets, and use the test set as the query images. In order to evaluate retrieval results on the iNaturalist dataset, we create a custom query and gallery set: We follow the original train/validation split, using all 579K training images to train the encoder model, and use 20 percent of the validation dataset (17K images) as the query set and the rest (78K images) as the gallery set. 

In addition to the baselines from the classification experiment, we include triplet loss \cite{Weinberger2009triplet, schroff2015facenet} and FashionNet \cite{Liu2016deepfashion} results as well, as they are more appropriate for retrieval tasks. We were unable to find equivalent results, or an official implementation for Ge \etal \cite{ge2018deep} (a different encoder network is used in the paper), and hence do not include their results here. In Figure \ref{fig:fullretreival}, we show results of the proposed three losses versus the baselines on DeepFashion In-Shop dataset, iNaturalist and Modelnet40. Our losses are clearly superior to the baseline results, with HiMulConE showing greater improvement at smaller $k$. 

Moreover, a recent study \cite{musgrave2020metric} showed that top-k retrieved results have flaws, and proposed a metric, Mean Average Precision at R (MAP). We report results in Table \ref{table:retrieval_map}. 

\begin{table}
\begin{center}
\begin{adjustbox}{width=0.9\columnwidth,center}
\begin{tabular}{ c|c|c|c|c } 
 \toprule[2pt]
    \multicolumn{1}{c|}{} & \multicolumn{2}{|c|}{DeepFashion} & \multicolumn{2}{|c}{ModelNet40} \\
 \toprule[1.5pt]
  & Category & Product & Category & Product\\ 
  & NMI &  NMI & NMI &  NMI \\
   \toprule[1.5pt]
 SimCLR & 0.15 & 0.73 & 0.31 & 0.52  \\

 CE & 0.1 & 0.66 & 0.12 & 0.4 \\ 
 
 SupCon & 0.57 & 0.68 & 0.57 & 0.69 \\ 
 
 HiMulCon & 0.57 & 0.8 & \textbf{0.62} & \textbf{0.88} \\ 
 
 HiConE & 0.58 & 0.78 & 0.61 & \textbf{0.88} \\ 
 
  HiMulConE & \textbf{0.59} & \textbf{0.81} & \textbf{0.62} & \textbf{0.88} \\ 
 \bottomrule[2pt]
\end{tabular}
\end{adjustbox}
\end{center}
\caption{NMI on DeepFashion In-Shop and ModelNet40. CE represents cross entropy. Product NMI represents its mean NMI.}
\label{table:clustering}
\end{table}

\begin{table}
\caption{Retrieval results using MAP evaluation metrics.}
\begin{center}
\begin{adjustbox}{width=0.9\columnwidth,center}
\begin{tabular}{ c|c|c|c } 
\toprule[2pt]
  & DeepFashion & iNaturalist & ModelNet40 \\ 
 \toprule[2pt]
 SupCon  & 31.5 & 61.5 & 21.6 \\ 

 HiMulConE  & \textbf{35.6} & \textbf{66.9}  & \textbf{26.0} \\ 
 \bottomrule[2pt]
\end{tabular}
\end{adjustbox}
\end{center}

\label{table:retrieval_map}
\end{table}

\begin{figure}[t]
\begin{minipage}{0.5\linewidth}
\includegraphics[width=1\linewidth]{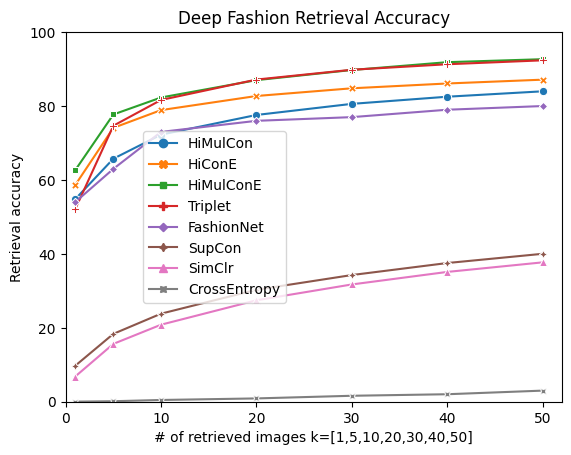}
\end{minipage}%
\hfill
\begin{minipage}{0.5\linewidth}
\includegraphics[width=1\linewidth]{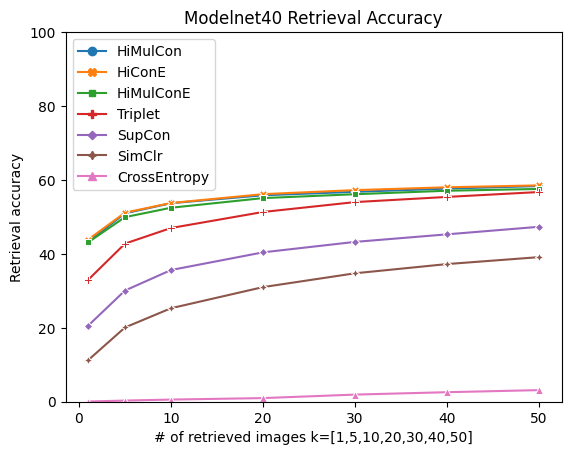}
\end{minipage}%
\caption{Retrieval results on the unseen splits of DeepFashion and ModelNet. HiMulConE performs better than the other approaches, especially at lower $k$. }
\label{fig:fashionretrieval}
\end{figure}


\subsection{Generalizability  to Unseen Data}
\subsubsection{Setup}
In order to evaluate the performance of our models on unseen data, we split DeepFashion and Modelnet40 datasets into a seen and unseen set. Each dataset is further split into train, validation and test sets. The seen set is used to train the encoder network, and the model is frozen. For the classification task on unseen data, a classifer is trained on the embeddings generated by the encoder network that was trained on the seen data, the encoder network is not finetuned on unseen data.

\vspace{-8pt}
\subsubsection{Classification}
We finetuned the pretrained model on the seen dataset. To obtain results in the unseen dataset, we only train the classifier on the embeddings generated from the encoder network that was trained on the seen dataset. Table \ref{table:classification_top1} shows the top-1 accuracy of classification accuracy on DeepFashion In-Shop and ModelNet40. It is seen that the proposed methods obtain better results than the baselines on the unseen part of both datasets, while obtaining comparable results to SupCon on the seen part. In theory, HiConE exploits the properties of the dataset by respecting the semantic relationships of neighboring levels, penalizing pairs by magnitudes that are correlated with the distance between the pair in the embedding space. HiMulCon on the other hand is a fixed penalty based on level differences, penalizing pairs on the distance between them in the label space. A hybrid of the two methods, HiMulConE can leverage differences from both the label and the embedding spaces. In Table \ref{table:classification_top1} we see that HiConE does better on the dataset with significant semantic overlap in different levels of the tree (DeepFashion), and the gap is much smaller in the semantically well separated dataset (ModelNet).

\vspace{-5pt}
\subsubsection{Image Retrieval}
The experimental setup is similar to section \ref{sec:retrieval}, with the difference being that the data used comes from the unseen set only. The results are presented in Figure \ref{fig:fashionretrieval}. Once again, our losses perform better than the baselines, particularly at lower values of $k$. This shows the generalisability of our work in comparison to the baselines. Since our approach can incorporate the label hierarchy in the network loss, the embedding space preserves the label-space hierarchy.

\subsubsection{Clustering}
Clustering is another downstream task that can be used to evaluate the quality of the embeddings. As in Ho \etal \cite{ho2020exploit}, we use K-means and the NMI \cite{vinh2010information} score to evaluate clustering quality. We first generate the embeddings for all the images in the unseen test set, and perform K-means in the representation space. Clustering is done at two levels: the lowest (category) and highest (product ID) levels in the tree. At the category level, K is set to the number of categories in the dataset, and NMI measures the consistency between the category labels and \textit{clusterId}. At the ID level, for each cateogry, we perform K-means, with K set to the number of products in that category. The mean of ID-level NMIs, across all categories, is reported in the Product NMI columns in Table \ref{table:clustering}. The significant improvement over the baseline in product NMI shows that our approach maintains separability for sub-categories within a category, and also shows that our approach preserves the hierarchical relationship between labels in the representation space.

\section{Limitations and Social Impact}
A limitation of our approach is the requirement of hierarchical labels for learning the encoder network, which can be expensive to acquire. In addition, our approach has been tested on datasets where labels have a tree-like structure, where each node only has one parent. However, it is relatively straightforward to extend this to general graph structures. Another common limitation arises from the underlying data used for experiments. Biases in the data \cite{steed2021image} can be learned by the model, which can have significant societal impact. Explicit measures to debias data through re-annotation or restructuring the dataset for adequate representation is necessary.

\section{Conclusion}
\label{sec:conclusion}
Hierarchically categorized data is common in the real world, and our novel approach provides a general framework for utilizing all available label data, reducing to standard supervised or self-supervised approaches in the absence of sufficient data. Our approach generalizes well on a variety of downstream tasks and unseen data, and significantly outperforms the evaluated baselines. In future work, we would like to extend this work to multi-label scenarios that are not in a hierarchical framework, and to other modalities and multi-modal settings, incorporating modalities such as speech and language. 

{\small
\bibliographystyle{ieee_fullname}
\bibliography{egbib}
}

\end{document}


\title{Use All The Labels: A Hierarchical Multi-Label Contrastive Learning Framework: Supplemental Material}  

\maketitle
\thispagestyle{empty}
\appendix

\section{Details of Seen and Unseen Data}
In the DeepFashion dataset, there are 25,900 training images, 12,612 validation images and 14,218 test images, where we use query images as test images in the task of category classification. We first train our model on seen categories, and finetune the classifier on unseen categories for the task of category classification. For the task of image retrieval, we apply the features from the header to calculate the feature distances between a query image and gallery images. Note that there is no overlap in categories between seen and unseen data, and there is no overlap in images in train and test sets.

ModelNet40 is a synthetic dataset of 3,183 CAD models from 40 object classes, and we split the data into 22 seen and 18 unseen categories. In the seen categories, the numbers of training, validation, and test images are 16,896, 4224, and 5,280, while in the unseen categories, the numbers of them are 13,662, 3,414, and 4,320. For the image retrieval task, the gallery dataset has 11,221 images and the query has 6,017. Since there is no official retrieval split, we split it similar to the validation/test ratio in DeepFashion In-Shop.

\section{Qualitative assessment of Image Retrieval on Unseen data}
\begin{figure*}[t]
\begin{center}
\includegraphics[width=\textwidth]{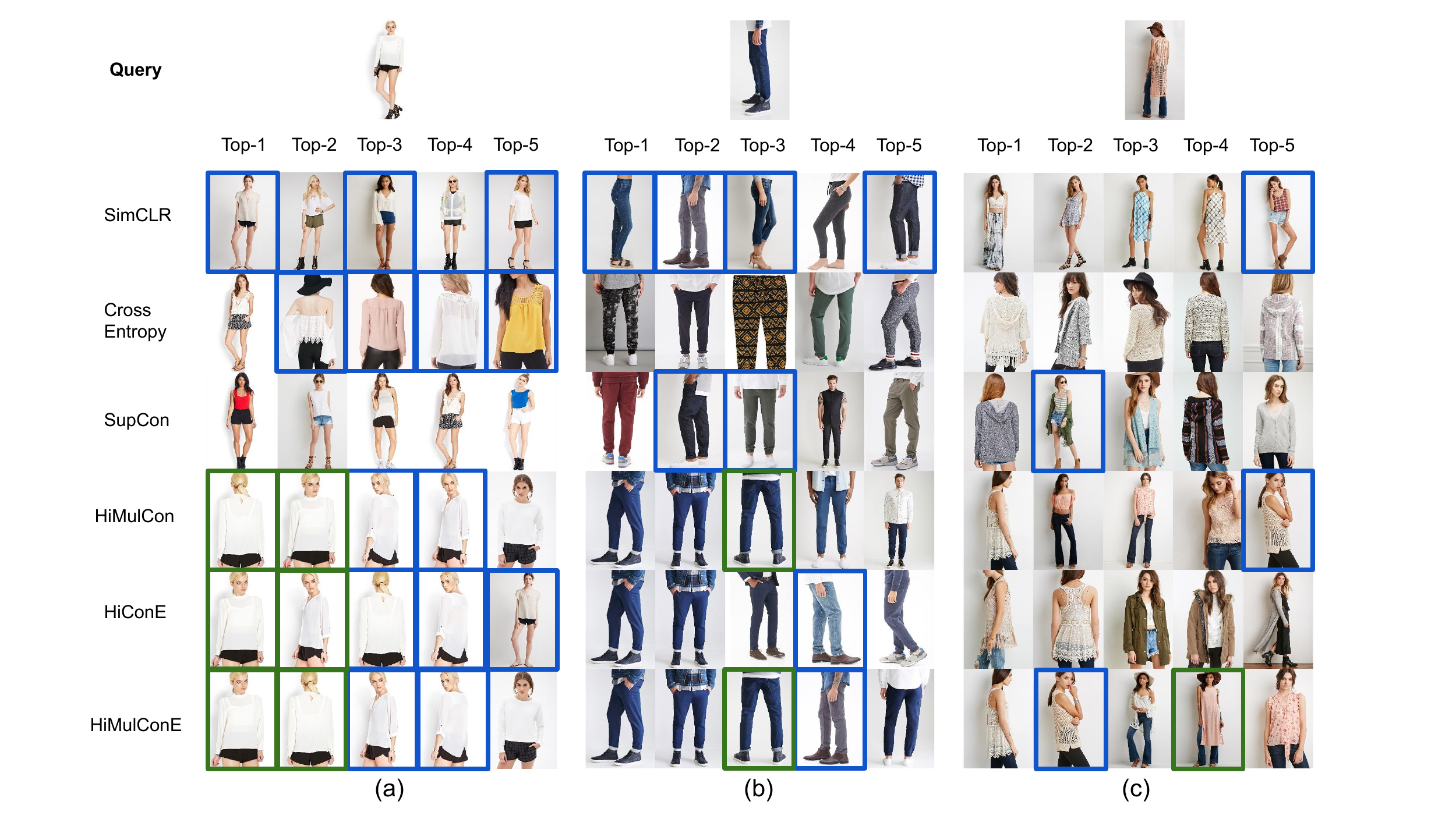}
\end{center}
\caption{Retrieval visualizations on DeepFashion In-Shop Dataset. The top row has 3 query images, and the rows below show top-5 results of the six methods. Green bounding boxes represent correctly retrieved results, and the blue bounding boxes represent the correct category but wrong product ID.}
\label{fig:retrievalresult1}
\end{figure*}

In Figure \ref{fig:retrievalresult1}, we show a visualization of the retrieved top-5 images by different algorithms. The top row has 3 query images, and the other rows show their corresponding retrieved top-5 results. The green bounding boxes represent correct retrieved images, and the blue bounding boxes represent wrongly retrieved images but in the correct categories. Recall that the image retreival task is setup at the product level in the hierarchy, which is higher in depth in comparison to the category level. In Figure \ref{fig:retrievalresult1}(a), the top-2 retrieved images of the three proposed algorithms are both correct. Although SimCLR and the cross entropy loss do not retrieve correct images, most retrieved images obtain correct categories. In the more challenging example in Figure \ref{fig:retrievalresult1}(b), the retrieved images of SimCLR have the best number of correct categories (4), but the corresponding product IDs are not correct among all the 5 retrieved images. In contrast, the proposed HiMulCon and the HiMulConE can both retrieve correct images (top-3). Considering the fact that denims are very similar to pants and the fact that some denims look very similar to each other, e.g. the retrieved images from our proposed algorithms look very similar, our proposed losses have a powerful ability to distinguish similar products. The query image in Figure \ref{fig:retrievalresult1}(c) is the most challenging image in these three examples, as it has both tees and denims. We can see that only HiMulConE retrieves the correct image, while all other methods, including the two individual losses that we propose, fail to find the correct product ID. Comparing the results of the proposed three losses to the three baselines, we can see that most retrieved images of our algorithms return a tee-denim combination, which is a reasonable context given the query image. We argue that the combined loss HiMulConE leads to the best learning ability among all methods, with the model showing good separability at both the category and sub-category levels.

\section{Ablation Study}
\label{sec:ablation}

\subsection{Sampling Strategy}
The sampling approach described in main paper tries to sample at least one positive pair from each level in the hierarchy. This strategy ensures that each level has postive pairs, and it also guarantees hard negative mining as the pairs from lower levels can be considered to be hard samples for higher levels. This strategy becomes more relevant with an unbalanced tree structure, as random sampling from a skewed tree structure can lead to the network overfitting to sub-trees with higher image density. For instance, the ratio of image count in the largest and the smallest categories in Deep Fashion training set is over $30$. In a statistical study, we found that the random sampling strategy would result in no positive pairs (other than augmented versions of the same image) in over $20\%$ of batches. 

We measure the efficacy of our hierarchical batch sampling strategy by comparing its performance with a completely random strategy and a sampling strategy that only ensures multiple positive pairs at the category level. The experiments were all performed with the DeepFashion dataset, with the HiConE loss. All hyperparameters are kept constant throughout this set of experiments. Table \ref{table:ablation_sampling} shows the results, a completely random sampling approach results in a significant deterioration in category prediction.

\begin{table}
\begin{center}
\begin{tabular}{ c|c|c } 
\toprule[2pt]
  Approach & Seen & Unseen \\ 
  \toprule[2pt]
 Hierarchical batch sampling & \textbf{80.52} &  \textbf{75.29}  \\
  
 Category level sampling & 79.81 & 72.63 \\ 
 
 Random Sampling & 77.96 & 71.59 \\
 \bottomrule[2pt]
\end{tabular}
\end{center}
\caption{Ablation study on effectiveness of the batch sampler}

\label{table:ablation_sampling}
\end{table}


\subsection{Layer Penalty in HiMulCon}
\label{sec:ablation_lambda}
The guiding intuition in designing the penalty term in HiMulCon is that higher level pairs need to be forced closer than lower level pairs in the hierarchy. To that end, we evaluate various functions for $\lambda_l= F(l)$, where the functions have a proportional relationship to the level. The functions can also be replaced with an ordered list of penalty values, which can be be treated as tunable hyperparameters, but we leave that for future work.

\begin{table*}
\begin{center}
\begin{tabular}{ c|c|c|c|c|c} 
\toprule[2pt]
  $f(l)$ & $exp(l)$ & $exp(\frac{1}{\left | L \right | - l})$ & $2^{l}$ & $2^{\frac{1}{\left | L \right | - l}}$ & $\frac{1}{\left | L \right | - l}$ \\ 
  \toprule[2pt]
  DeepFashion &73.12 & 74.29 & 73.6 & 73.8 & 73.47 \\
  
  ImageNet &77.94 & 79.14 & 78.36 & 78.22 & 78.15 \\
 \bottomrule[2pt]
\end{tabular}
\end{center}
\caption{Study of various candidates for $\lambda_l$ for HiMulCon}
\label{table:ablation_lambda}
\end{table*}

We evaluate the performance of category prediction on the unseen data validation set of Deep Fashion and the whole validation set of ImageNet for various $f(l)$, and $exp(\frac{1}{l})$ is the candidate picked for other experiments. Keeping with our intuition, note that all of the functions described in Table \ref{table:ablation_lambda} have an directly proportional relationship with level $l$. We also performed sanity check experiments where we evaluated various functions that had a inversely proportional relationship as well, their performance was lower than those seen in the table.

\subsection{Layer Penalty in HiMulCon: Sanity check experiments}
In addition to the ablation study conducted to evaluate different $\lambda_l$ candidates listed in Table \ref{table:ablation_lambda}, we also conducted some sanity check experiments to verify correctness of the implementation and to validate our intuitive understanding of the effect of the penalty term on the learned representations. Similar to the generalisability experiments in the main paper, all the results reported in Table \ref{table:ablation_lambda_supp} are conducted on the validation set of the unseen data split in the DeepFashion InShop dataset \cite{Liu2016deepfashion}. In the ablation study covered in Section \ref{sec:ablation_lambda}, we limited our study to functions that were directly proportional to the level. This corresponded to the intuition that the closer the lowest common ancestor of the image pair is to the leaf node, the higher the pair level is, and the higher the penalty should be. In this study, we present experiments with functions that are inversely proportional to the layer level.

First, we evaluated the Identity function $\lambda_l =  (\mathbb{1})$, whose effect would be that all layers would be penalized to the same amount, and all layer pairs would be equivalent positive pairs.This function would reduce the HiMulCon loss to become approximately equal to the SupCon \cite{khosla2020supervised} loss, but HiMulConE would still benefit from the hierarchical constraint. 

Next, we study a collection of exponential functions that decrease with increasing level. The performance significantly deteriorates with increasing proportional penalty terms, validating the relationship between label structure and the loss penalty term.

\section{t-sne visualization}
\begin{figure*}
    \begin{minipage}{0.33\linewidth}
    \includegraphics[width=1\linewidth]{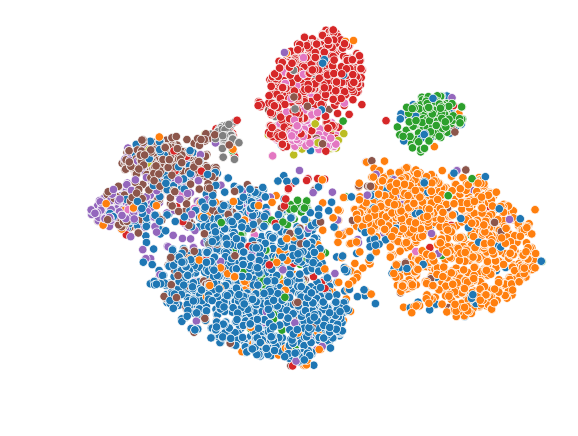}
    \captionof*{figure}{HiMulConE}
    \end{minipage}%
    \hfill
    \begin{minipage}{0.33\linewidth}
    \includegraphics[width=1\linewidth]{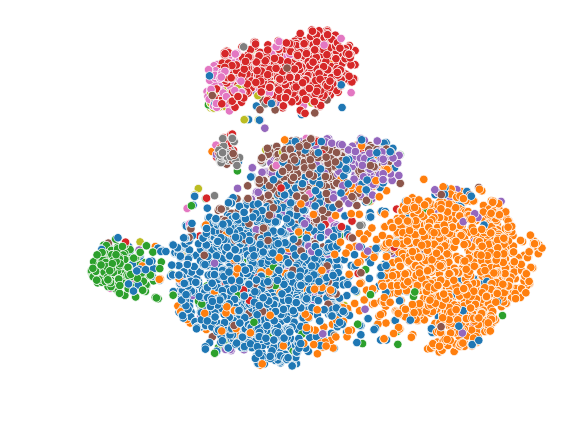}
    \captionof*{figure}{HiMulCon}
    \end{minipage}%
    \hfill
    \begin{minipage}{0.33\linewidth}
    \includegraphics[width=1\linewidth]{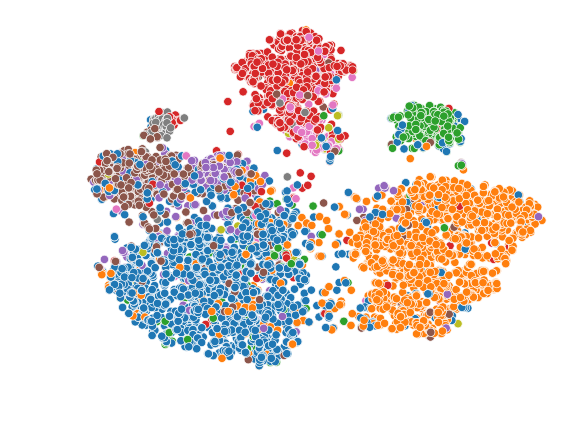}\hfill
    \captionof*{figure}{HiConE}
    \end{minipage}%
    \hfill
    \\[\smallskipamount]
    \begin{minipage}{0.33\linewidth}
    \includegraphics[width=1\linewidth]{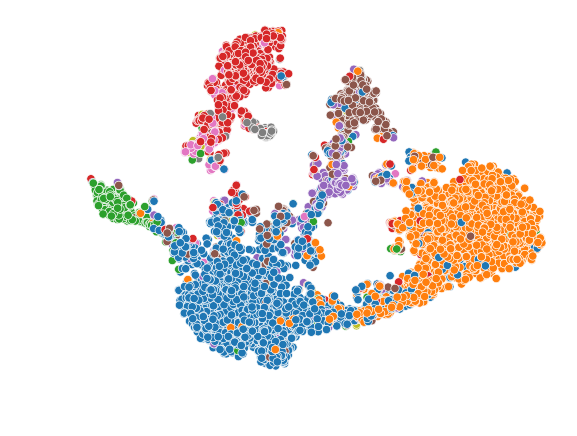}\hfill
    \captionof*{figure}{SupCon}
    \end{minipage}%
    \hfill
    \begin{minipage}{0.33\linewidth}
    \includegraphics[width=1\linewidth]{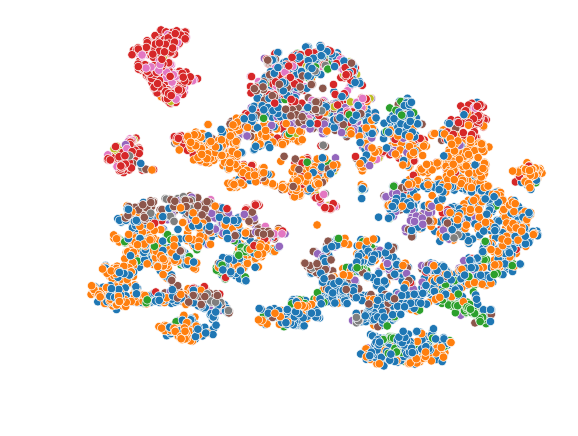}\hfill
    \captionof*{figure}{SimCLR}
    \end{minipage}%
    \hfill
    \begin{minipage}{0.33\linewidth}
    \includegraphics[width=1\linewidth]{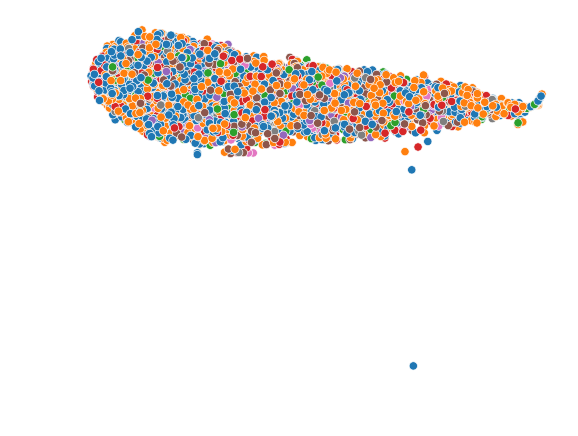}\hfill
    \captionof*{figure}{Cross Entropy}
    \end{minipage}%
    \begin{picture}(0,0)
        \put(-30,-20){\includegraphics[height=1.5cm]{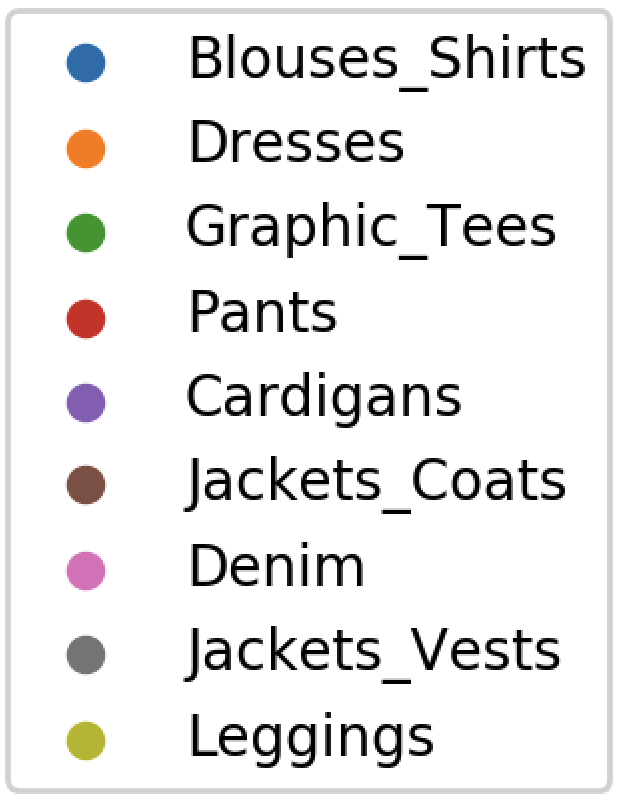}}
    \end{picture}
    \hfill
    \\[\smallskipamount]
    \caption{t-sne visualizations of the Deep Fashion dataset}\label{fig:tsne}
\end{figure*}

We project the test image embeddings into 2 dimensions through t-sne \cite{van2008visualizing} and visualize the results on Deep Fashion dataset in Figure \ref{fig:tsne}. The three proposed losses have a clear category level separability. Interestingly, the semantically similar categories, like \textit{Pants} and \textit{Denim}, as well as \textit{Cardigans} and \textit{Jacket\_Coats} are much closer to each other in the embedding space compared to unrelated categories. Although SimCLR and SupCon have some clusters, this is not correlated with category labels, and there is significant mixing of different categories in the clusters from those approaches.

\begin{figure*}
\begin{center}
    \begin{minipage}{0.33\textwidth}
    \includegraphics[width=1\linewidth]{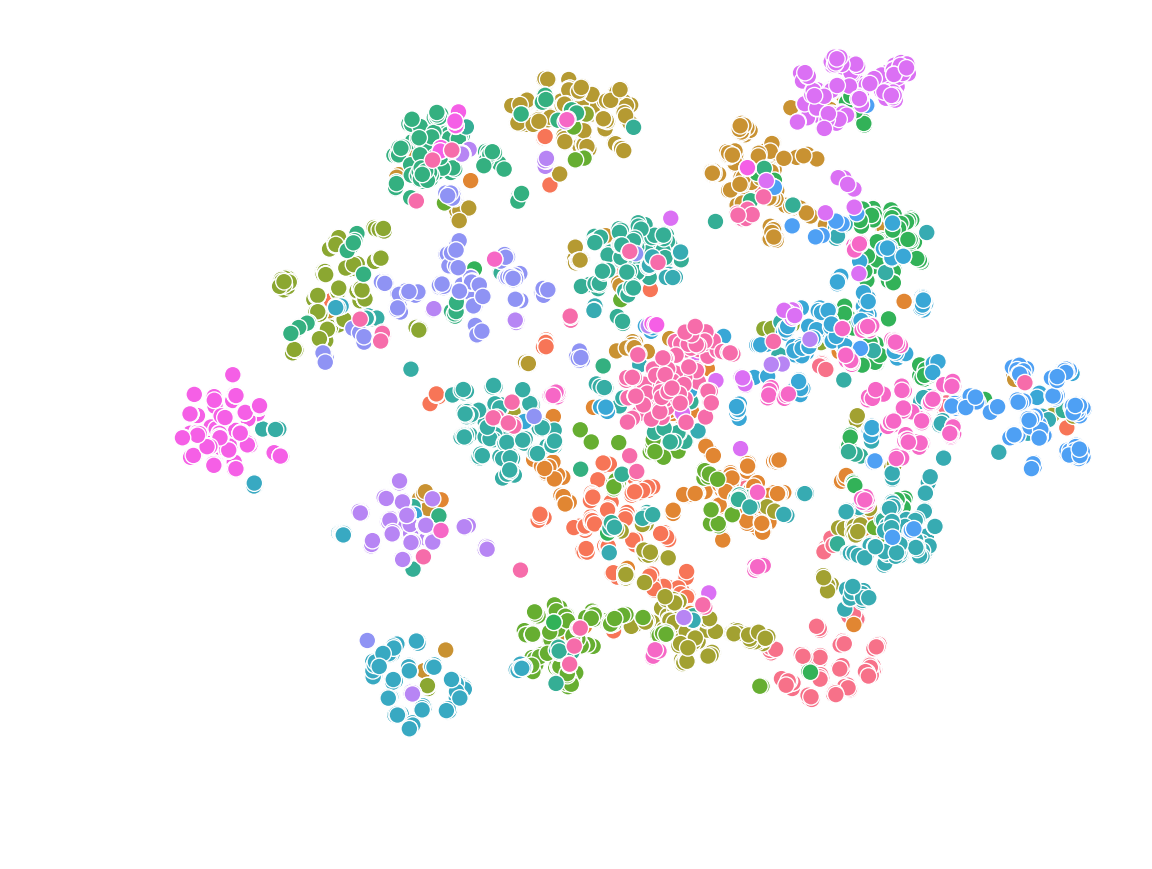}
    \captionof*{figure}{HiMulConE}
    \end{minipage}%
    \begin{minipage}{0.33\textwidth}
    \includegraphics[width=1\linewidth]{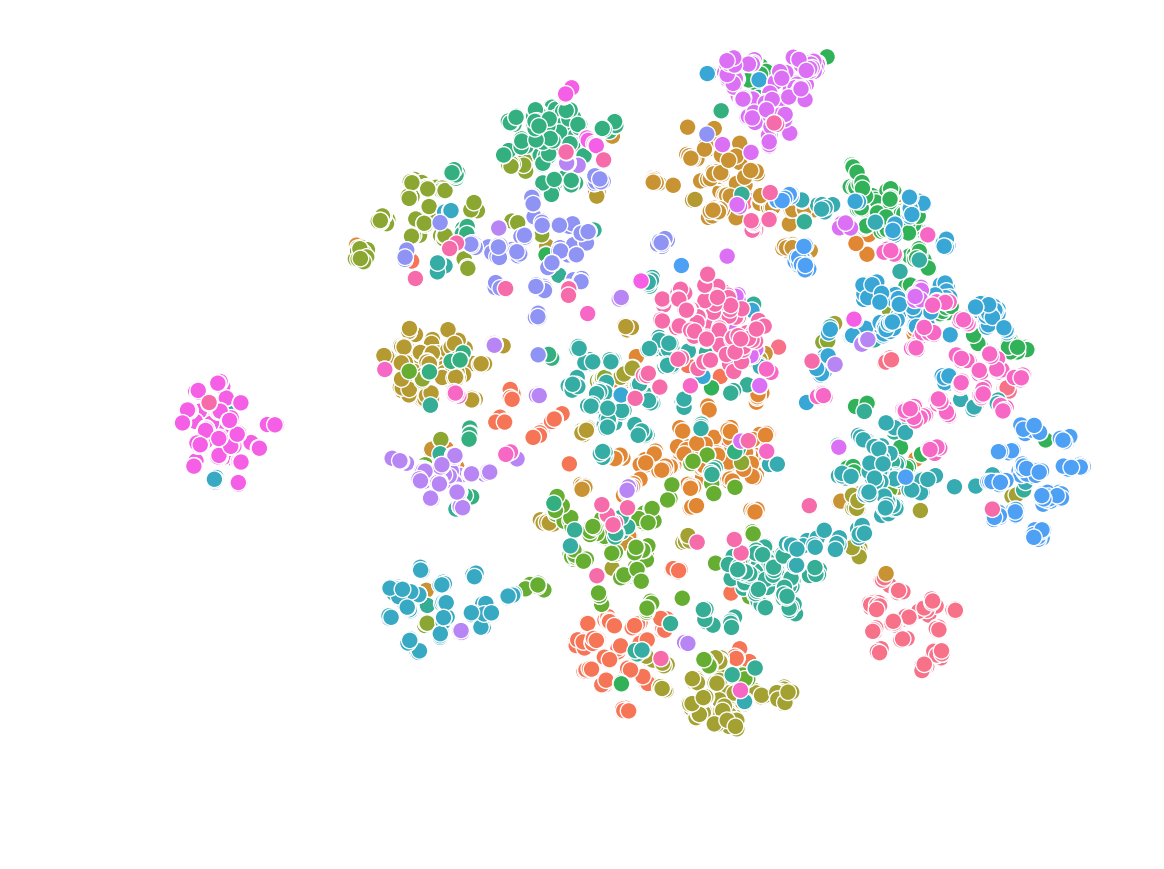}
    \captionof*{figure}{HiMulCon}
    \end{minipage}%
    \begin{minipage}{0.33\textwidth}
    \includegraphics[width=1\linewidth]{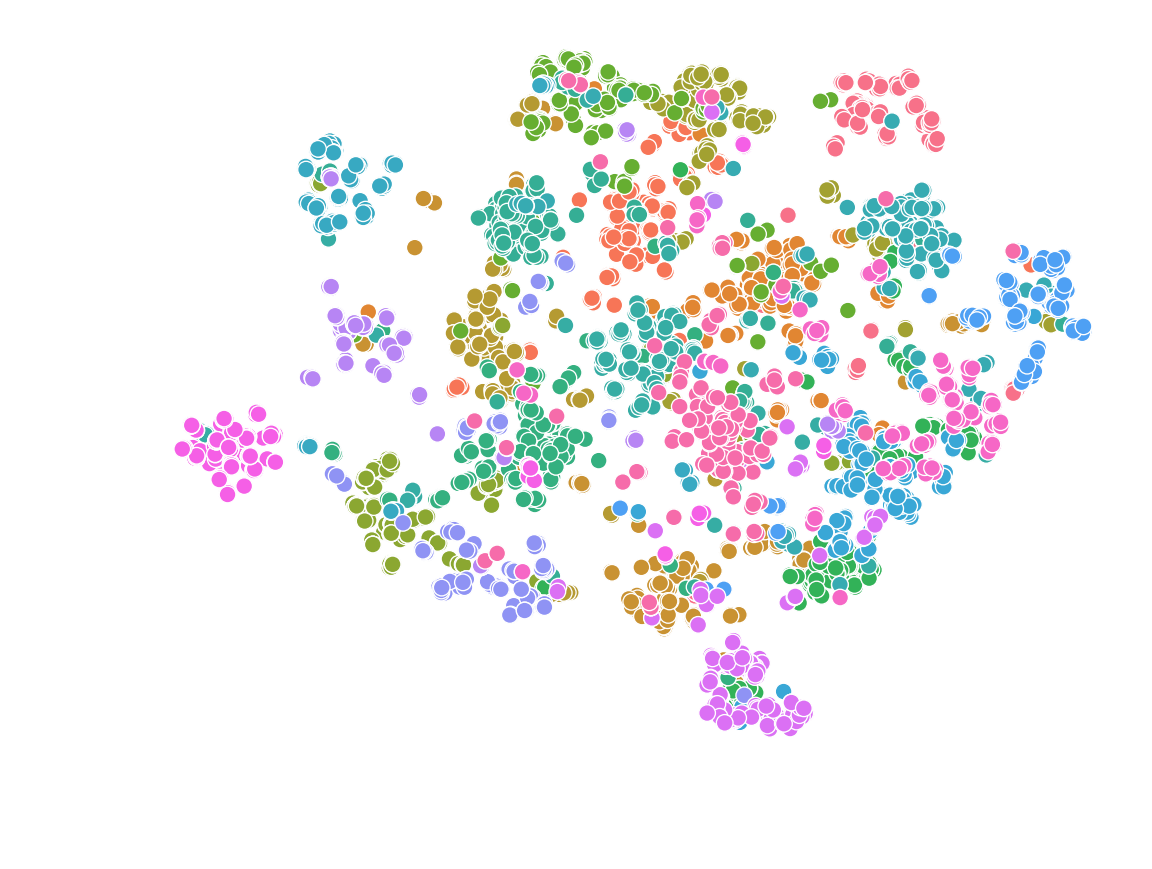}\hfill
    \captionof*{figure}{HiConE}
    \end{minipage}%
    \hfill
    \\[\smallskipamount]
    \begin{minipage}{0.33\textwidth}
    \includegraphics[width=1\linewidth]{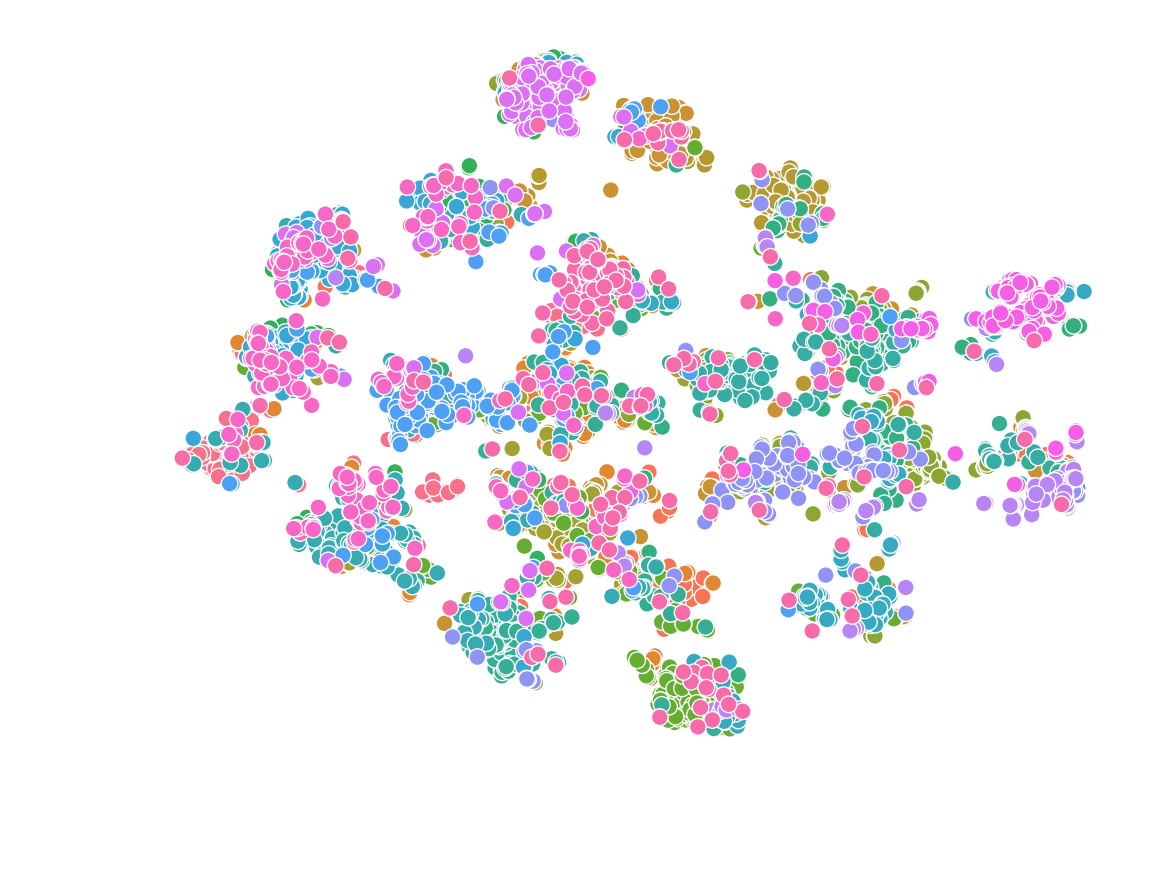}\hfill
    \captionof*{figure}{SupCon}
    \end{minipage}%
    \begin{minipage}{0.33\textwidth}
    \includegraphics[width=1\linewidth]{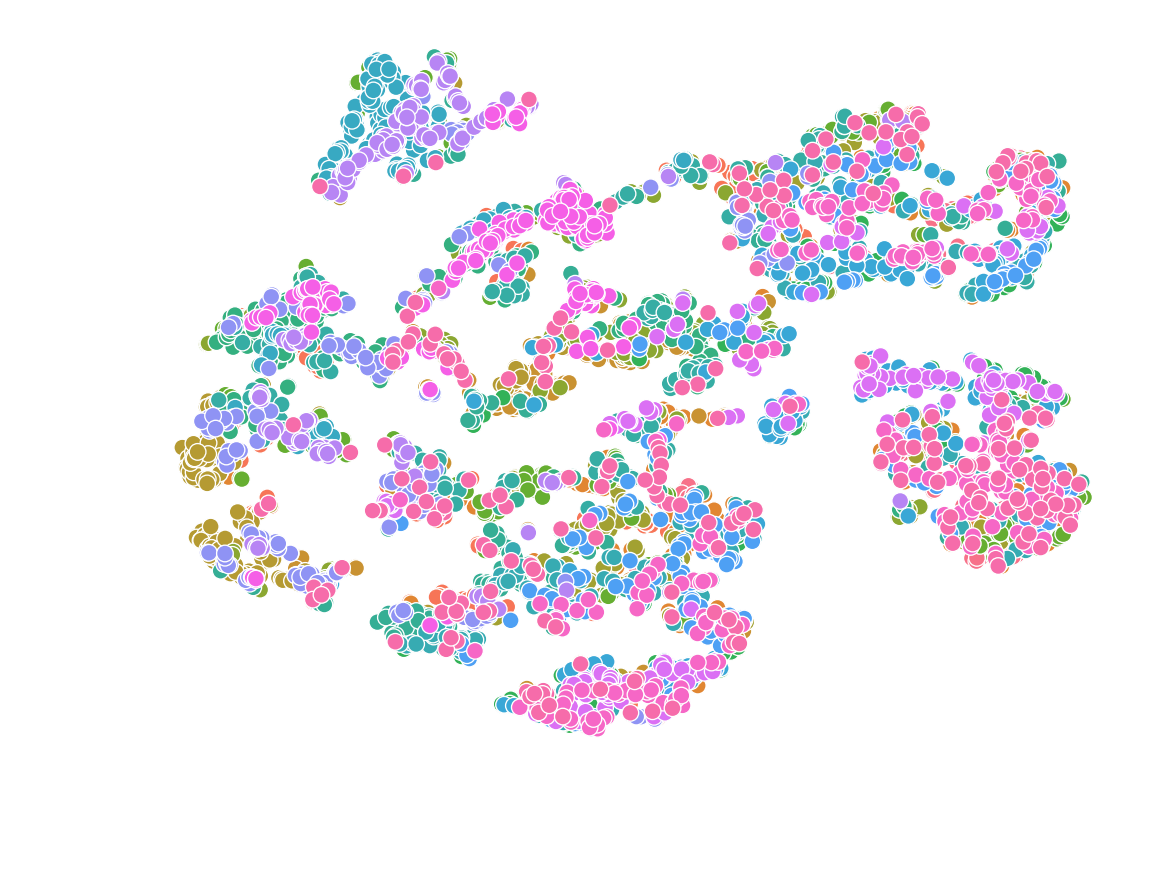}\hfill
    \captionof*{figure}{SimCLR}
    \end{minipage}%
    \begin{minipage}{0.33\textwidth}
    \includegraphics[width=1\linewidth]{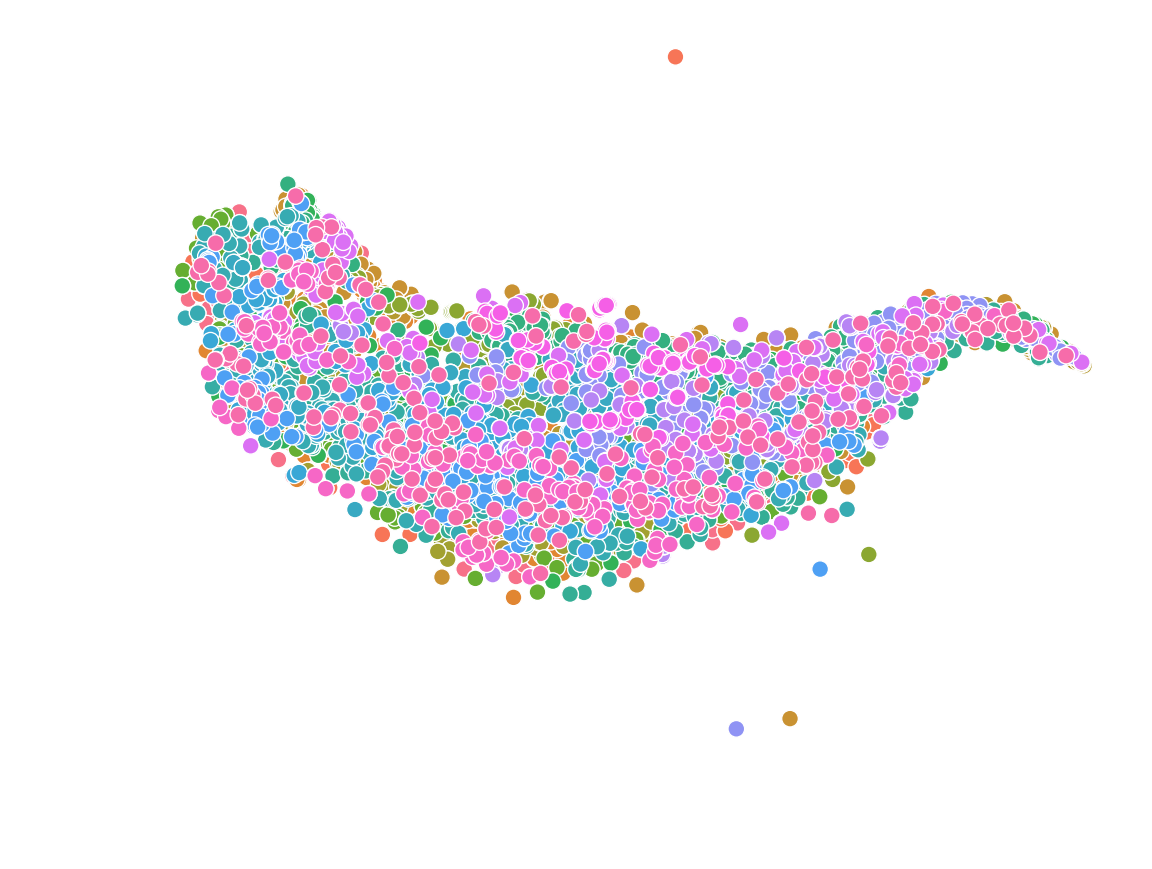}\hfill
    \captionof*{figure}{Cross Entropy}
    \end{minipage}%
    \begin{picture}(0,0)
        \put(-30,35){\includegraphics[height=4cm]{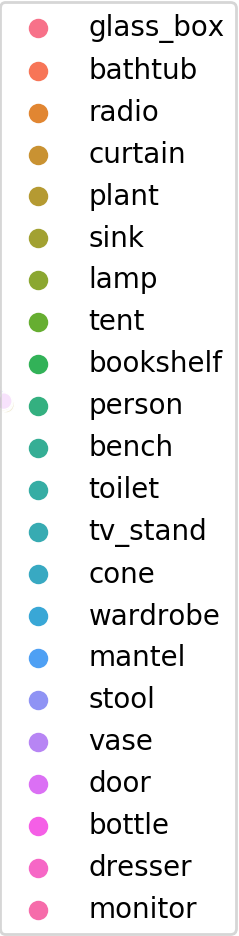}}
    \end{picture}
    \hfill
    \\[\smallskipamount]
    \caption{t-sne visualizations of the Modelnet40 dataset.  approaches.}\label{fig:tsne_modelnet}
    \end{center}
\end{figure*}

In Figure \ref{fig:tsne_modelnet}, we present Modelnet40's t-sne visualizations. Consistent with our findings in quantitative analysis in Table 2 of the main paper, we achieve good separability with our approaches. Both SimClr and SupCon have clear sub-spaces in the representation, but they have a poorer correlation with category labels

\begin{figure}
\includegraphics[height=0.3\textheight,width=\linewidth]{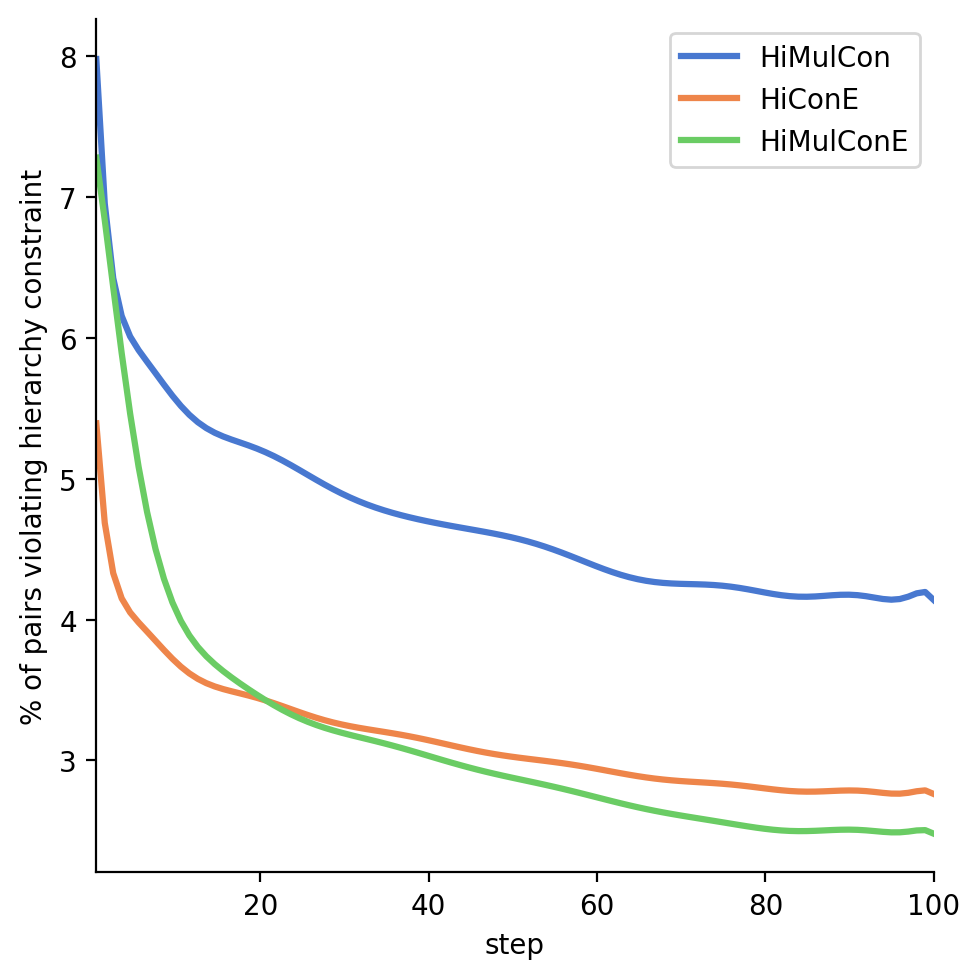}
\caption{Hierarchical Violations during training}
\label{fig:hc}
\end{figure}
\begin{table}
\begin{center}
\begin{tabular}{ c|c}
\toprule[2pt]
Approach & Classification \\ &  Accuracy \\
  \toprule[2pt]
  HiMulCon, $\lambda_l=\mathbb{1}$ & 73.1 \\
  
  HiMulConE, $\lambda_l=\mathbb{1}$ & 73.8 \\
  
 HiMulCon, $\lambda_l=\frac{1}{l})$ & 70.8 \\
 
  HiMulCon, $\lambda_l=exp(\frac{1}{l}))$ & 70.29 \\
 
   HiMulCon, $\lambda_l=2^\frac{1}{l})$ & 69.4 \\
 \toprule[2pt]
  SupCon & 73.6 \\
  
  HiMulCon, $\lambda_l=exp(\frac{1}{\left | L \right | - l})$ & 74.29 \\
 
  HiMulConE, $\lambda_l=exp(\frac{1}{\left | L \right | - l})$ & \textbf{76.07} \\
 \bottomrule[2pt]
\end{tabular}
\end{center}
\caption{Study of various candidates for $\lambda_l$. The last three rows are the results with the validation set on the approaches used in the main paper.}
\label{table:ablation_lambda_supp}
\end{table}
\begin{table}
\begin{center}
\begin{tabular}{ c|c} 
\toprule[2pt]
   Loss Function & \% of pairs \\ & violating constraint  \\
  \toprule[2pt]
  HiMulCon & 7.45 \\

  HiConE & 4.26 \\
  
 HiMulConE & \textbf{3.74} \\
  
  SupCon & 9.4 \\
  
  SimClr & 14.95 \\
 
  Cross Entropy & 27.43 \\
 \bottomrule[2pt]
\end{tabular}
\end{center}
\caption{Hierarchy violations in the test set. Lower numbers indicate fewer pairs violated the hierarchical constraint.}
\label{table:hc}
\end{table}

\section{Hierarchy Constraint Study}
We defined the hierarchy constraint as the requirement that the loss between image pairs from a higher level in the hierarchy will never be higher than the loss between pairs from a lower level. The hierarchy constraint is violated if the lower level pair has a lower loss than a higher level. In figure \ref{fig:hc}, we track the frequency of hierarchy violations during the training process on our unseen split of DeepFashion. This value is tracked before the loss is calculated, and is aggregated across all batches in an epoch. All three losses reduce the frequency of constraint violations, but since HiMulCon does not directly optimize for the hierarchy constraint, it does poorly in comparison to the other two losses. Additionally, the penalty term defined by $\lambda_l$ in HiMulConE helps reduce the frequency of the violations better than HiMulCon. 

Next, we studied the frequency of hierarchy violations on the held out test set. We constructed pairs at random, obtained the lowest common ancestor for each pair, computed the distance between the image pairs, and tracked the frequency of hierarchy constraint violation in the embedding space. In the distance based computation, a violation is when a pair from a higher level will be closer than a pair from a lower level. Table \ref{table:hc} presents the hierarchical constraint violation occurrence on the test set.
The reduction in hierarchical constraints with convergence, and in the held out validation set, is additional evidence that the representation learning framework presented in this paper preserves the hierarchical label structure in embedding space.


{\small
\bibliographystyle{ieee_fullname}
\bibliography{egbib}
}